\documentclass[letterpaper, 10 pt, conference]{ieeeconf}  

\IEEEoverridecommandlockouts                              

\usepackage{graphics} 
\usepackage{epsfig} 
\usepackage{mathptmx} 
\usepackage{times} 
\usepackage{amsmath} 
\usepackage{amssymb}  
\usepackage{multirow}
\usepackage{booktabs}
\usepackage{ulem} 
\usepackage{cite}
\usepackage{stfloats}
\usepackage{xcolor}
\usepackage{makecell}
\usepackage[ruled,vlined,linesnumbered]{algorithm2e} 

\usepackage[caption=false]{subfig}
\usepackage{hyperref}
\hypersetup{hidelinks,
	colorlinks=true,
	allcolors=black,
	pdfstartview=Fit,
	breaklinks=true}

\title{\LARGE \bf
Exploring Bottlenecks in VLM-LLM Navigation: How 3D Scene Understanding Capability Impacts Zero-Shot VLN}

\author{
 Ziyi Xia$^{1,\dagger}$, Chaoran Xiong$^{1,\dagger}$,~\IEEEmembership{Graduate~Student~Member,~IEEE}, Litao Wei$^{1}$, \\ Xinhao Hu$^{1}$, and Ling Pei$^{1,2,\ast}$,~\IEEEmembership{Senior~Member,~IEEE}
\thanks{$^{\dagger}$Ziyi Xia and Chaoran Xiong contribute equally to this work.}
\thanks{$^{\ast}$Corresponding Author: Ling Pei.}
\thanks{$^1$The authors are with the Shanghai Key Laboratory of Navigation and Location Based Services, Shanghai Jiao Tong University (SJTU), Shanghai 200240, China. $^2$Ling Pei is also with the State Key Laboratory of Submarine Geoscience, SJTU. (e-mail: \{matcha.latte; sjtu4742986; oscar0731; xinhaohu1203; ling.pei\}@sjtu.edu.cn).}
}

\begin{document}

\maketitle
\thispagestyle{empty}
\pagestyle{empty}

\begin{abstract}
Zero-shot vision-and-language navigation (VLN) has gained significant attention due to its minimal data collection costs and inherent generalization. This paradigm is typically driven by the integration of pre-trained Vision-Language Models (VLMs) and Large Language Models (LLMs), where VLMs construct 3D scene graphs while LLMs handle high-level reasoning and decision-making. However, a critical bottleneck exists in this system: current 3D perception models prioritize pixel-level accuracy, directly conflicting with the strict computational limits and real-time efficiency demanded by embodied navigation. To address this gap, this paper quantifies the actual impact of 3D scene understanding capability on VLN performance. Based on typical VLM-LLM frameworks, we propose statistical success rate (SR) upper bounds for two core subsystems: 1) the slow LLM planner, which relies on topological mapping semantics, and 2) the fast reactive navigator, which utilizes spatial coordinates and bounding boxes to execute LLM decisions. Evaluations using state-of-the-art 3D scene understanding models validate our proposed bounds and reveal a perception saturation phenomenon, indicating that improvements in perception accuracy beyond a certain threshold yield diminishing returns in navigation success. Our findings suggest that 3D scene understanding for VLN should pivot away from strict pixel-level precision, prioritizing instead navigation-relevant core vocabularies and accurate bounding box proportions.
\end{abstract}

\section{INTRODUCTION}
\label{sec:intro}
Vision-and-Language Navigation (VLN) is a fundamental embodied artificial intelligence task in which an agent navigates to a target location by perceiving visual observations and reasoning over natural language instructions\cite{SSM}. Recently, zero-shot approaches have emerged as a promising paradigm for VLN due to their training-free nature, high deployment flexibility, and strong generalization ability.

Existing zero-shot Vision-Language Model and Large Language Model (VLM-LLM) pipelines typically process visual inputs with a VLM at every navigation step and then query the LLM to directly output actions. This architecture can be viewed as an all-slow-brain paradigm. However, such methods suffer from substantial computational overhead, including high token consumption and low temporal efficiency. Recent slow-fast brain collaborative frameworks have effectively alleviated these computational bottlenecks. Therefore, we adopt the SFCo-Nav \cite{SFcoNav} framework as the baseline for our VLN methodology.

\begin{figure}
    \centering
    \includegraphics[width=\columnwidth]{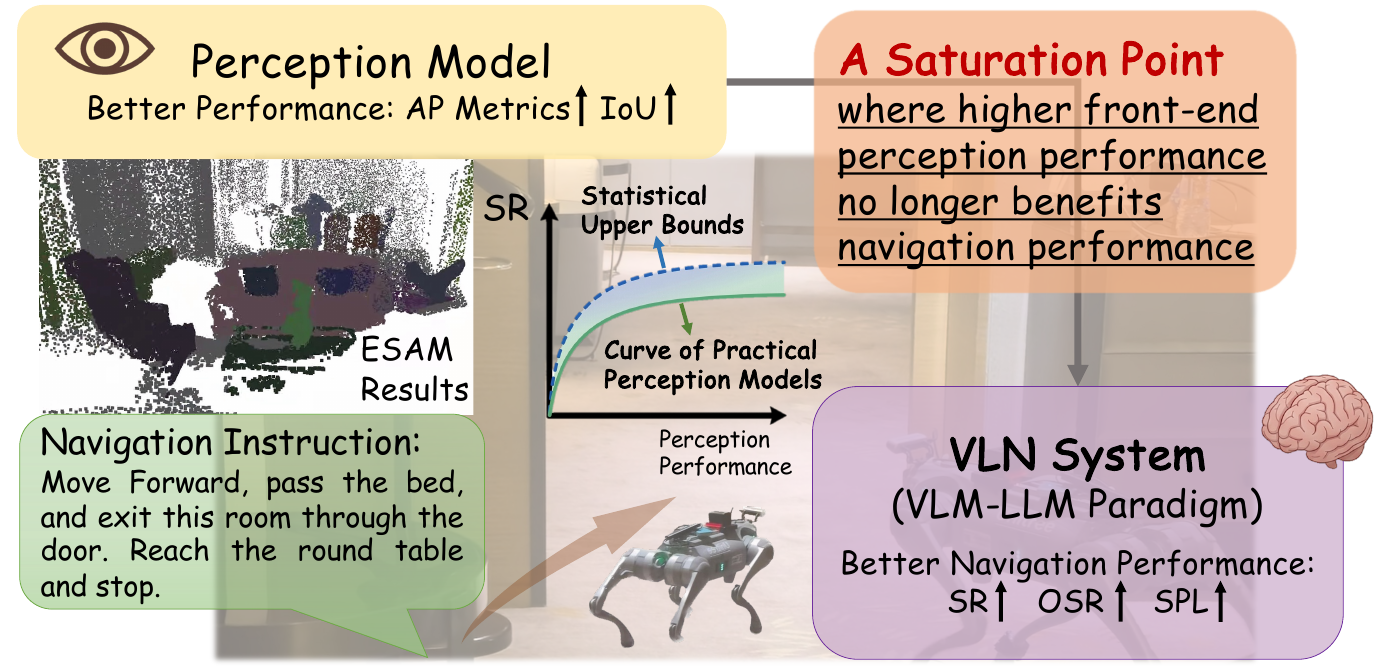}
    \caption{We study how 3D scene understanding capability affects zero-shot VLN in typical VLM-LLM frameworks, and derive statistical SR upper bounds for the slow LLM planner and the fast reactive navigator which is validated by experiments with state-of-the-art perception models.}
    \label{fig:abstract}
    \vspace{-0.6cm}
\end{figure}

In traditional all-slow-brain paradigms, the front-end perception module mainly relies on Visual Foundation Models (VFMs). For example, NavCoT \cite{NavCoT} uses BLIP \cite{BLIP} to translate visual observations into textual descriptions. Similarly, NavGPT \cite{NavGPT} employs BLIP-2 \cite{BLIP2} to generate descriptions for 24 different views at each viewpoint, which are then concatenated and summarized by GPT \cite{GPT} into a single sentence. This process is a cross-modal translation from visual signals to natural language, which inevitably introduces information loss. Although MapGPT \cite{MapGPT} avoids textual translation by feeding raw visual observations of all navigable viewpoints directly into GPT-4V \cite{GPT} at each step, it incurs higher token usage and inference latency.

\begin{figure*}[t]
    \centering
    \includegraphics[width=0.8\linewidth]{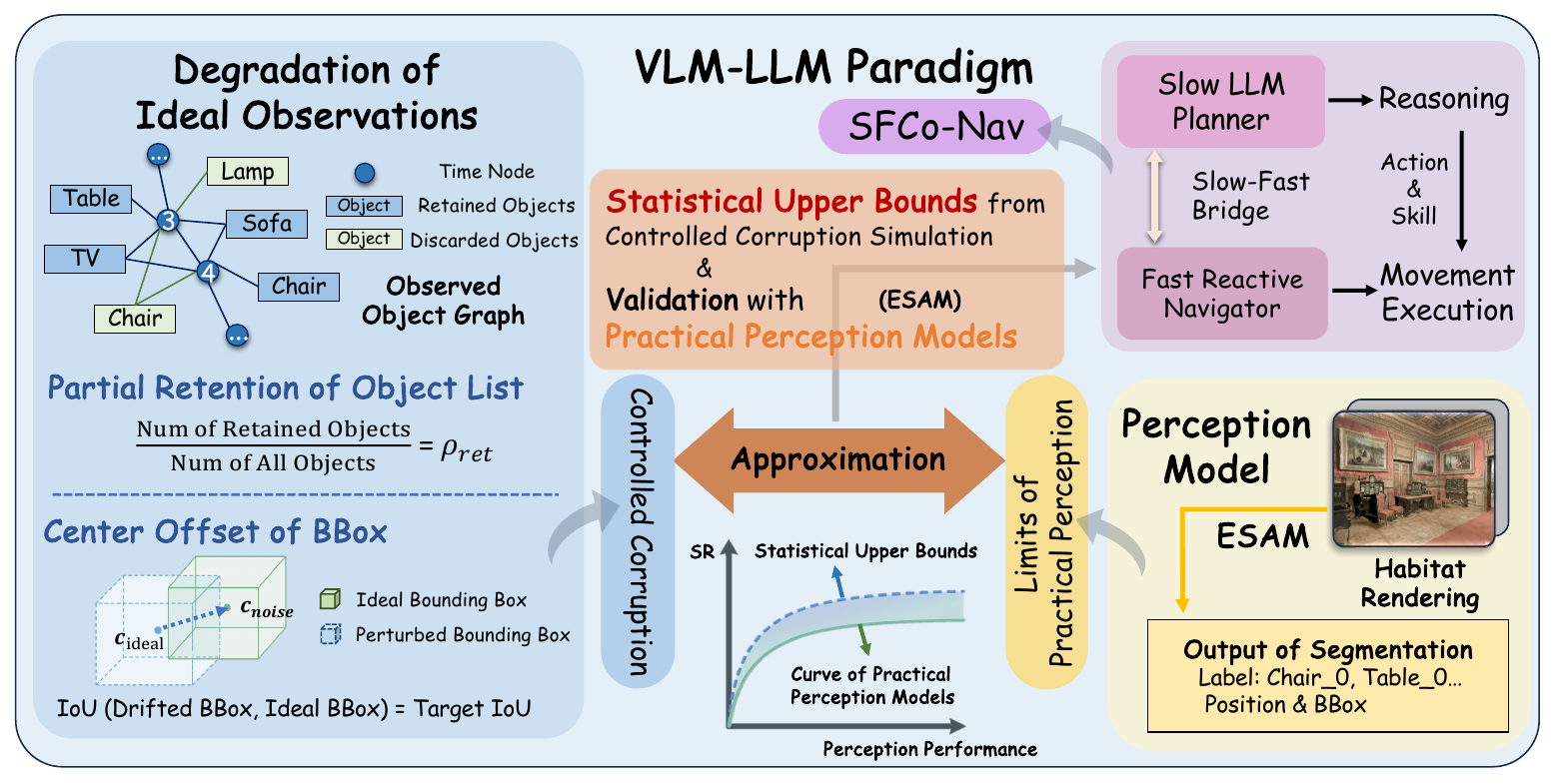}
    \caption{
    Overview of our analysis framework. We quantify how 3D scene understanding affects zero-shot VLN by studying two core modules in a typical VLM-LLM pipeline. For the slow LLM planner, we simulate missed detections with the retention ratio $\rho_{ret}$ and quantify semantic information retained in the degraded topological graph relative to the ideal graph using the matching score $S_{match}$. For the fast reactive navigator, we simulate controlled bounding box drift with the target IoU $\phi_{iou}$ to measure the effect of localization drift on action execution. By analyzing the relationship between $\rho_{ret}$, $\phi_{iou}$, and the navigation success rate (SR), we reveal statistical upper bounds for both modules and show a clear perception saturation effect.
    }
    \label{fig:overview}
    \vspace{-0.5cm}
\end{figure*}

As VLN tasks gradually shift from discrete environment to continuous environment, agents require not only semantic understanding but also spatial coordinates and bounding boxes. Prior all-slow-brain architectures have attempted to incorporate spatial information. For example, Open-Nav \cite{OpenNav} extracts object coordinates and semantic labels using the spatial-understanding VLM SpatialBot \cite{SpatialBot} and the fine-grained object recognition model RAM \cite{RAM}, while NavGPT identifies bounding boxes with Fast-RCNN \cite{Fast-RCNN}.

Previous VLN methods typically selected perception modules that were state of the art at the time, with model choice driven primarily by standalone perception capability. However, much less attention has been paid to whether these modules are suitable for lightweight online embodied navigation, as well as to their computational cost and latency. More importantly, prior work has rarely examined how much perception quality is actually required to achieve a target navigation success rate. In practice, navigation is a multi-module system-level task. Given a desired success rate, it is necessary to consider how perception quality, computation, and latency should be allocated and traded off across different modules under limited resource budgets. 

From the perspective of Pareto optimality, a system is not optimized by maximizing a single metric in isolation, but by operating at a point where further improvement in one aspect would require disproportionate cost while bringing only limited system-level benefit. In zero-shot VLN, this implies that perception should not be scaled up solely to improve standalone accuracy. Instead, it should be evaluated according to how much it contributes to downstream navigation success under the overall computation and latency budget.

Under this perspective, it is important to estimate the navigation performance supported by different levels of perception capability and to examine whether the return from improved perception exhibits saturation. Accordingly, as shown in \ref{fig:overview}, we design targeted degradation strategies for the two core components of the current VLM-LLM navigation architecture, namely the slow LLM planner and the fast reactive navigator, while focusing only on the most influential observation factors and evaluating them on a moderate number of navigation instructions. Based on this, we estimate statistical upper bounds and validate them using the state-of-the-art Embodied SAM (ESAM) \cite{esam}, which leverages the Segment Anything Model (SAM) for real-time online 3D instance segmentation. We use statistical upper bound to denote an empirically estimated performance ceiling over sampled navigation instructions, rather than a formally proven probabilistic bound. Through these experiments, we aim to explore the potential of using target navigation performance to guide the capability requirements of the front-end perception module.
Our contributions can be summarized as follows:
\begin{enumerate}
\item A statistical navigation success rate upper bound analysis of the impact of 3D scene understanding capability on zero-shot VLN. We derive statistical upper bounds on navigation success rate for both the slow LLM planner and the fast reactive navigator, and reveal a clear perception saturation effect in both modules.

\item A simulation framework that models two key perception factors for the slow LLM planner and fast reactive navigator, enabling statistical upper-bound estimation. For the former, the framework models missed detections with the retention ratio and semantic graph quality with the matching score. For the latter, it simulates object localization noise through bounding box center perturbations calibrated to different overlap ratios relative to the ideal boxes.

\item An ESAM-based validation study that grounds the proposed upper-bound analysis in practical 3D perception models. Results across ESAM variants with different capability levels align well with the predicted upper bounds, confirm the saturation effect, and indicate that the remaining gap is mainly due to false positives and distorted bounding box aspect ratios. We further show that reliable recognition of a small navigation-relevant core vocabulary is sufficient for satisfactory navigation performance.
\end{enumerate}

\section{METHODS}
\label{sec:methods}
The current VLM-LLM paradigm consists of two main components, namely the slow LLM planner and the fast reactive navigator, which correspond to the Slow Brain and Fast Brain in the SFCo-Nav system, respectively. These two components place different demands on perceptual information. Accordingly, we design separate perception degradation strategies for each component.

For the slow LLM planner, we primarily investigate the impact of missed detections on LLM reasoning and decision making. Specifically, we sample different retention ratios to partially mask the ideal semantic observation. For the fast reactive navigator, we focus on how bounding box localization noise affects low-level motion execution. Thus, we modify the navigation skills so that the next viewpoint is computed from bounding box geometry. We then perturb the centers of ideal bounding boxes with offsets chosen to match specified target IoU levels, thereby simulating different degrees of localization noise.

\subsection{Ideal Observations Setup}
In navigation, not all perceived objects are equally important. We therefore tend to preserve in the ideal observations those key objects that are explicitly mentioned in the instruction, as well as objects that help the agent understand the current scene. For a given viewpoint $v$, let $\mathcal{O}^{v}_{gt} = (o_1, o_2, \dots, o_N)$ denote the ideal sequence with fixed ordering. Each object $o_i$ is associated with a ideal bounding box $b^{v}_{i, gt}$ and a correct semantic label. This sequence serves as the ideal observation for a certain viewpoint, representing perfect perception.

\subsection{Slow LLM Planner: Missed Detections}
\begin{figure}
    \centering
    \includegraphics[width=1.0\linewidth]{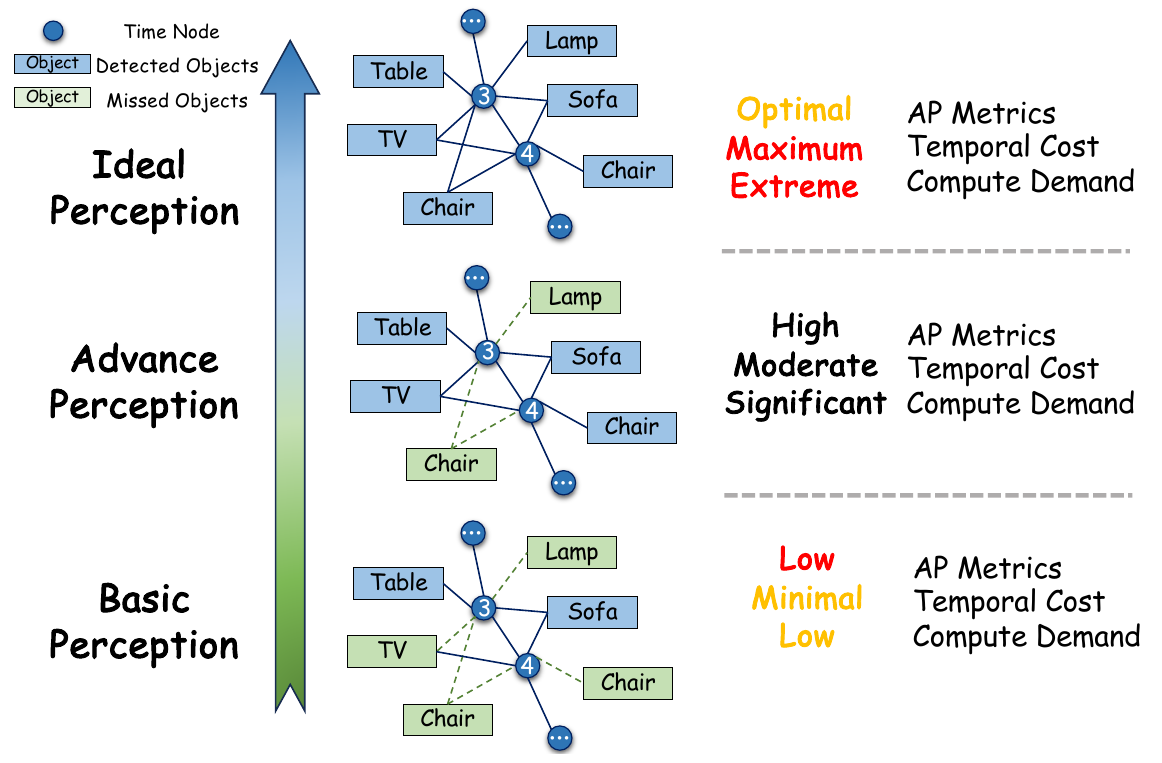}
    \caption{Topological graphs and efficiency trade-offs under different perception performance.}
    \label{fig:slow}
    \vspace{-0.8cm}
\end{figure}

In SFCo-Nav, the Slow Brain constructs a semantic topological mapping graph, as shown in \ref{fig:slow}, represented as a temporal-object bipartite graph $\mathcal{G} = (\mathcal{T}, \mathcal{V}_{obj}, \mathcal{E})$, where $\mathcal{T}$ represents time nodes of the trajectory, $\mathcal{V}_{obj}$ represents semantic objects, and edges $(t, o) \in \mathcal{E}$ indicate that object $o$ was observed at time $t$. 

To simulate missed detections, we use retention ratio $\rho_{ret} \in [0, 1]$ to obtain the truncated observation $\mathcal{O}^{v}_{slow} = \{o_1, \dots, o_K\}$ where $K = \lfloor \rho_{ret} \cdot N \rceil$. Fixed ordering ensures the nested-subset property: 
\begin{equation}
    \rho_1 < \rho_2 \implies \mathcal{O}(\rho_1) \subset \mathcal{O}(\rho_2).
\end{equation}

Upon completion of a trajectory, we compare the degraded graph $\mathcal{G}_{deg}$ with the ideal graph $\mathcal{G}_{ideal}$ using two alignment metrics to quantify topological integrity:

\begin{enumerate}
    \item \textbf{Object Recall Score ($S_{obj}$)}: Measures the fraction of distinct object names preserved under degraded perception compared with the ideal observation set. This reflects the agent's overall semantic coverage of the environment.
    \item \textbf{Temporal-Semantic Edge Score ($S_{edge}$)}: Measures the proportion of correctly preserved $(t, o)$ associations. This focuses on the temporal consistency and the agent's ability to re-identify objects during movement.
\end{enumerate}

The final matching score is defined as a weighted sum: $S_{match} = \lambda S_{edge} + (1-\lambda) S_{obj}$. We then perform a bucket-based analysis by binning $S_{match}$ and calculating the SR within each bin to reveal the mathematical correlation between graph integrity and navigation success. The precise calculation of these scores is detailed in Algorithm 1.

For each discrete $\rho_{ret}$, we aggregate the results across all evaluated trajectories to compute the corresponding SR and establish the relationship among the retention ratio $\rho_{ret}$, the topological integrity $S_{match}$, and the navigation performance SR. Specifically, we analyze the correlation in three stages: (i) $\rho_{ret}$ vs. SR, to illustrate how semantic observation completeness affects navigation Success Rate; (ii) $\rho_{ret}$ vs. $S_{match}$, to quantify how missed detections degrade the graph structure; and (iii) $S_{match}$ vs. SR, to reveal the minimum topological information required for successful reasoning. This multi-stage analysis provides detailed insight into how the decision-making process of the slow LLM planner responds to varying degrees of semantic information loss.

\begin{algorithm}[h]
\caption{Calculation of Matching Score}
\label{alg:matching_score}
\SetKwInput{KwIn}{Input}
\SetKwInput{KwOut}{Output}
\SetKwProg{Fn}{Function}{ :}{}

\KwIn{Degraded graph $\mathcal{G}_{deg}$, Ideal graph $\mathcal{G}_{ideal}$, Weight $\lambda$}
\KwOut{matching score $S_{match} \in [0, 1]$}

\BlankLine
    \BlankLine
    \tcp{Semantic Node Retrieval} 
    $\mathcal{V}_{obj}^{ideal} \leftarrow \{o.\mathrm{name} \mid o \in \mathcal{V}_{obj, ideal}\}$\;
    $\mathcal{V}_{obj}^{deg} \leftarrow \{o.\mathrm{name} \mid o \in \mathcal{V}_{obj, deg}\}$\;
    $S_{obj} \leftarrow \frac{|\mathcal{V}_{obj}^{deg} \cap \mathcal{V}_{obj}^{ideal}|}{|\mathcal{V}_{obj}^{ideal}|}$ 
    
    \BlankLine
    
    \tcp{Temporal-Semantic Edge Retrieval}
    $\mathcal{E}_{ideal} \leftarrow \{(t, o.\mathrm{name}) \mid (t, o) \in \mathcal{E}_{ideal}\}$\;
    $\mathcal{E}_{deg} \leftarrow \{(t, o.\mathrm{name}) \mid (t, o) \in \mathcal{E}_{deg}\}$\;
    $S_{edge} \leftarrow \frac{|\mathcal{E}_{deg} \cap \mathcal{E}_{ideal}|}{|\mathcal{E}_{ideal}|}$\;
    
    \BlankLine
    
    \tcp{Weighted Score Fusion}
    $S_{match} \leftarrow \lambda \cdot S_{obj} + (1-\lambda) \cdot S_{edge}$\;
    \KwRet $S_{match}$\;
\end{algorithm}
\vspace{-0.2cm}

\begin{figure}
    \centering
    \includegraphics[width=\columnwidth]{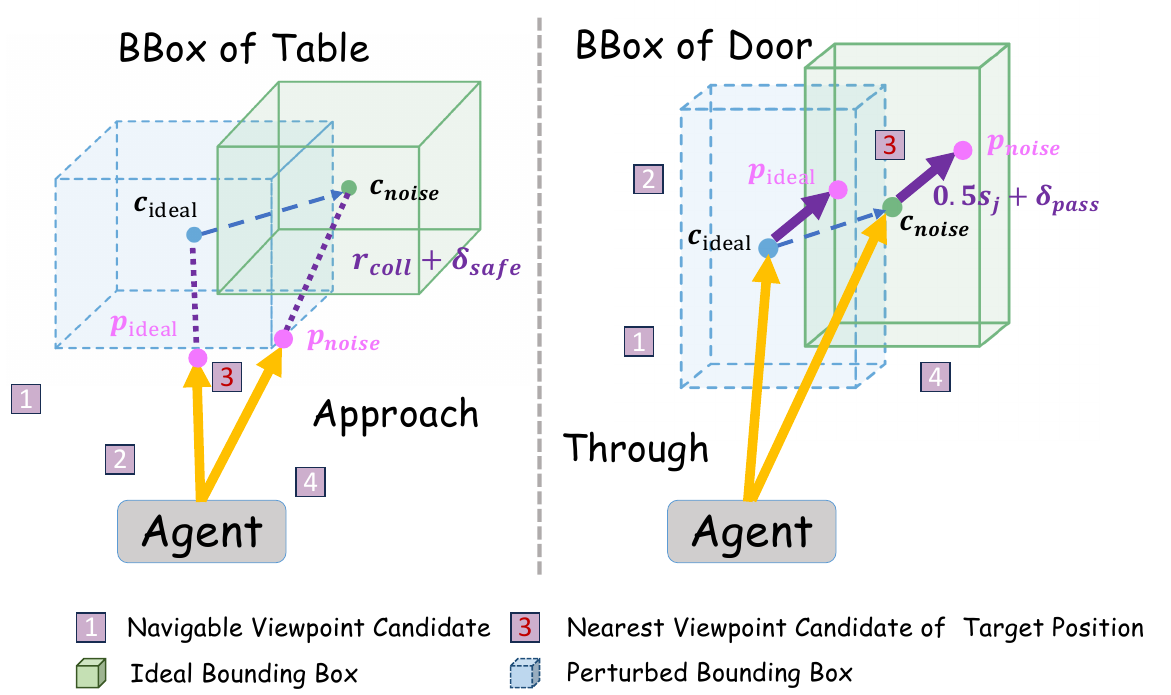}
    \caption{Center offsets to ideal BBox and updated skills.}
    \label{fig:fast}
    \vspace{-0.8cm}
\end{figure}

\subsection{Fast Reactive Navigator: Bounding Box Drift}
The Fast Brain of SFCo-Nav utilizes spatial coordinates for action execution. To simulate localization noise, we introduce target IoU $\phi_{iou} \in [0, 1]$. This variable represents the Intersection over Union (IoU) between the perturbed bounding box and the ideal bounding box for each object involved in the navigation task. By manipulating $\phi_{iou}$, we can uniformly scale the degree of spatial deviation across all observed objects. This metric directly reflects the spatial alignment between the perturbed bounding box and its ideal counterpart.  We use this protocol as a controlled probe of center localization drift, rather than as a complete model of all perception errors in real embodied systems.

For each object, we perturb the ideal bounding box $b_{ideal} = [\mathbf{c}_{ideal}, \mathbf{s}, \mathbf{R}]$ (representing centroid, dimensions, and rotation) to a noisy state $b_{noise}$. We maintain the original dimensions $\mathbf{s}=(w, h, d)$ and rotation $\mathbf{R}$, only applying a translation to the centroid:
\begin{equation}
    \mathbf{c}_{noise} = \mathbf{c}_{ideal} + \vec{\Delta},
\end{equation}
where $\vec{\Delta}$ is a displacement vector. To ensure sensor randomness, the direction of $\vec{\Delta}$ is chosen by randomly assigning $\pm 1$ to each axis, while the magnitude is scaled such that $\text{IoU}(b_{noise}, b_{ideal}) = \phi_{iou}$. This operation applies a uniform degree of spatial deviation across all observed objects. By manipulating $\phi_{iou}$, we examine the tolerance to translational drift and isolate the system's sensitivity to spatial noise. This design intentionally keeps scale and orientation fixed, and therefore does not model other practical perception failures such as scale estimation errors.

As depicted in \ref{fig:fast}, to approximate continuous-space execution within discrete environment, we upgrade movement execution with geometry-aware skills that leverage the perturbed bounding box $b_{noise}$ to determine the optimal viewpoint $v^*$:

    \subsubsection{Approach Skill} Designed for interaction targets (e.g., tables), this skill calculates a stopping point adjacent to the object. Let $\mathbf{p}_{agent}$ be the agent's current position and $\mathbf{u} = \frac{\mathbf{c}_{noise} - \mathbf{p}_{agent}}{\|\mathbf{c}_{noise} - \mathbf{p}_{agent}\|}$ be the unit vector pointing to the object. The target coordinate is defined as:
    \begin{equation}
        \mathbf{p}_{target} = \mathbf{c}_{noise} - \mathbf{u} \cdot (r_{coll} + \delta_{safe}),
    \end{equation}
    where $r_{coll}$ is the object's collision radius derived from dimensions $\mathbf{s}$, and $\delta_{safe}$ is a safety buffer.

    \subsubsection{Through Skill} Designed for traversable structures such as doors, this skill guides the agent to pass through the object by following its narrowest dimension. We determine the traversal direction by identifying the axis aligned with the object's minimal dimension $j = \arg \min(\mathbf{s})$ and retrieving the corresponding normal vector $\mathbf{n} = \mathbf{R}[:, j]$. To ensure the traversal logic aligns with the agent's forward momentum, we rectify the normal direction based on the incidence angle:
    \begin{equation}
        \mathbf{p}_{target} = \mathbf{c}_{noise} + \text{sgn}(\mathbf{u} \cdot \mathbf{n}) \mathbf{n} \cdot (0.5 \mathbf{s}_j + \delta_{pass}),
    \end{equation}
    where the sign function $\text{sgn}(\cdot)$ dynamically orients the traversal vector and $\mathbf{s}_j$ denotes the magnitude of the object's dimension along the traversal axis.

Finally, for both skills, the computed continuous coordinate $\mathbf{p}_{target}$ is projected onto the nearest navigable viewpoint: 
\begin{equation}
    v^* = \arg \min_{v \in \mathcal{V}_{nav}} \|\mathbf{p}_{target} - \text{pos}(v)\|.
\end{equation}

\begin{table}[t]
    \centering
    \caption{Perception Capability Evaluation for ESAM Variants.}
    \renewcommand\arraystretch{1.2}
    \belowrulesep=0pt
    \aboverulesep=0pt
    \resizebox{0.48\textwidth}{!}{
    \begin{tabular}{ccc|cccc}
    \toprule
    SV Train$^*$ & MV Train$^*$ & Total$^*$ & IoU & AP@25 & AP@50 & AP \\
    \cmidrule{1-7}
    300     & 0       & 300      & 6.3  & 6.2  & 4.1  & 2.1 \\
    600     & 0       & 600      & 12.5 & 11.0 & 6.0  & 2.8 \\
    80.10k  & 0       & 80.10k   & 46.1 & 48.2 & 32.3 & 18.6 \\
    80.10k  & 38.53k  & 118.63k  & 56.7 & 66.4 & 51.7 & 34.3 \\
    80.10k  & 38.53k  & 118.63k  & -    & 75.6 & 60.1 & 41.6 \\
    \bottomrule
    \end{tabular}}
    
    \vspace{2pt}
    \parbox{0.48\textwidth}{\footnotesize
    $^*$SV Train and MV Train denote the training iterations on single-view and multi-view data, respectively, and Total denotes their sum. The last row uses a 20-class vocabulary, while all others use 200 classes.}
    
    \label{tab:esam}
    \vspace{-0.6cm}
\end{table}

\begin{figure*}[!t]
    \centering
    \footnotesize
    \includegraphics[width=1.0\linewidth]{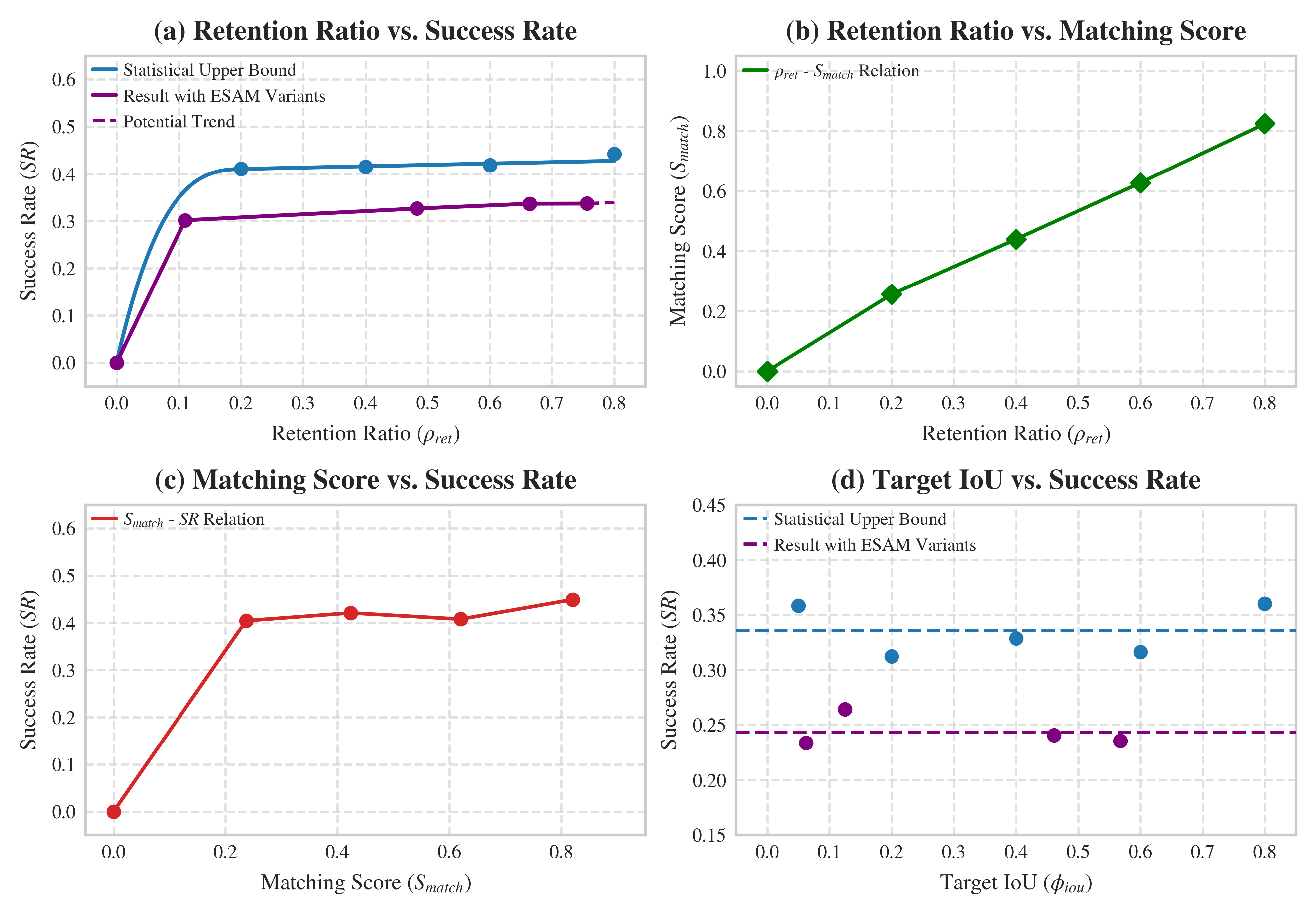}
    \caption{Impact of 3D scene understanding on navigation. Blue lines denote the statistical upper bounds of success rate (SR) under controlled corruption, and purple lines denote results of ESAM variants. From (a) to (c), $\rho_{ret}$ influences the intermediate matching score $S_{match}$ approximately linearly, while the mapping from $S_{match}$ to SR is saturating. Thus, SR becomes insensitive to further improvements in $\rho_{ret}$ beyond a certain range. The purple result is consistent with this upper bound trend. In (d), SR varies little with the control variable and the purple result again agrees with the upper bound behavior.}
    \label{fig:results}
    \vspace{-0.5cm}
\end{figure*}

\section{EXPERIMENTS}
Our experiments consist of two parts. In the first part, we conduct the statistical simulation described in Section~\ref{sec:methods}. In the second part, we select a practical perception model to validate the statistical upper bounds obtained. Considering the limitations of existing front-end perception modules in zero-shot VLN, and motivated by the increasingly strong spatial reasoning capabilities of LLMs together with the idea of topological map construction, we propose that semantic perception and fine-grained object localization can be unified within a Segment Anything Model (SAM)-based architecture. ESAM follows this SAM-based design philosophy while also meeting the real-time online requirements of embodied tasks. As a state-of-the-art perception model for embodied perception, it provides an effective testbed for validating the statistical upper bounds proposed in this work.
\subsection{Experimental Setup}
We evaluated our framework on the R2R dataset \cite{R2R}. For the statistical simulation experiments, we sampled approximately 600 instructions spanning 8 distinct scenes. For the evaluations integrating ESAM, each model was tested on approximately 300 instructions. The LLM agent of Slow Brain was powered by GPT-4o \cite{GPT} with the topological weighting parameter $\lambda$ set to 0.5. 

To emulate continuous observation of real-world deployment, we integrated ESAM into the Habitat simulator \cite{szot2021habitat}. At each viewpoint, the agent executes a 360-degree rotation, capturing 40 frames. To ensure semantic consistency, we use a semantic mapping that converts predicted vocabulary of ESAM into the corresponding object categories in R2R instructions. We use the model released by ESAM authors, with a 20-class vocabulary, as our best-performing perception model. 

To simulate degraded perception, we obtained four additional ESAM variants by partial training on ScanNet200 dataset\cite{scannet}. We evaluated all variants using AP metrics and IoU as shown in \ref{tab:esam}. We map the two simulation variables to measurable ESAM performance indicators. For the Slow Brain, $\rho_{ret}$ models whether object observations are retained in the semantic topological map. We therefore use $AP_{0.25}$ as its proxy, since the loose overlap threshold makes this metric primarily reflect missed-detection behavior rather than precise localization. For the Fast Brain, $\phi_{iou}$ models spatial perturbation, so we use instance-level box IoU as its counterpart. Specifically, for each ground-truth instance, we form an axis-aligned 3D bounding box from its associated point cloud, then compute the maximum IoU between this box and the predicted boxes of the same semantic class, and finally average this value over all instances. Therefore, these two metrics capture the same underlying factors as $\rho_{ret}$ and $\phi_{iou}$, respectively.

\subsection{Statistical Simulation Analysis}
For the Slow Brain, we investigated the impact of semantic completeness by varying $\rho_{ret} \in \{0.20, 0.40, 0.60, 0.80\}$, supplemented by a zero-recall case ($\rho_{ret}=0.00, SR=0\%$). For the Fast Brain, we varied target IoU $\phi_{iou} \in \{0.05, 0.20, 0.40, 0.60, 0.80\}$ to evaluate how its performance changes with different levels of localization noise.

As illustrated in Fig.\ref{fig:results}(a)-(c), we observe a clear saturation effect between $\rho_{ret}$ and SR. Once $\rho_{ret}$ reaches 0.20, SR quickly stabilizes at around 40\%. It remains nearly unchanged until $\rho_{ret}$ approaches 0.80, where SR shows only a slight improvement. However, this marginal gain is disproportionately small compared with the additional time and computational resources required to raise $\rho_{ret}$ from 0.20 to 0.80. We further observe a linear relationship between $\rho_{ret}$ and the matching score $S_{match}$, which in turn leads to a similar saturation trend between $S_{match}$ and SR. As shown in Fig.~\ref{fig:results}(d), once $\phi_{iou} > 0$, namely, once the predicted bounding box overlaps with the ground-truth bounding box, the SR remains nearly stable at around 33\% in the viewpoint-based VLN setting. This result corresponds to a statistical upper bound and indicates a clear saturation effect and we consider that the small fluctuations around this value are likely related to minor instability in LLM outputs.

Overall, these results suggest that the statistical simulation provides a reasonable upper bound on the impact of 3D scene understanding in both the slow LLM planner and the fast reactive navigator. In both cases, navigation performance shows a clear saturation trend as perception improves. This indicates that, once semantic retention or localization accuracy reaches a sufficient level, further improvements in perception are unlikely to yield substantial gains in SR.
\begin{table}[!t]
    \centering
    \caption{Impact of Perception Capability on Slow LLM Planner.}
    \renewcommand\arraystretch{1.2}
    \belowrulesep=0pt
    \aboverulesep=0pt
    \resizebox{0.4\textwidth}{!}{
    \begin{tabular}{c|c|ccc}
    \toprule
    Observation Source & $\rho_{ret}$ & SR & OSR & SPL \\
    \cmidrule{1-5}
    \multirow{4}{*}{Ideal Perception}
    & 0.20 & 41.0 & 49.9 & 35.2 \\
    & 0.40 & 41.5 & 51.2 & 35.9 \\
    & 0.60 & 41.8 & 51.2 & 35.8 \\
    & 0.80 & 44.3 & 54.8 & 38.6 \\
    \cmidrule(lr){1-5}
    \multirow{4}{*}{3D SAM Models}
    & 0.110 & 30.2 & 43.1 & 24.1 \\
    & 0.482 & 32.7 & 44.6 & 26.8 \\
    & 0.664 & 33.7 & 44.7 & 26.6 \\
    & 0.756 & 33.7 & 42.2 & 27.7 \\
    \bottomrule
    \end{tabular}}
    \label{tab:slow}
    \vspace{-0.25cm}
\end{table}
\begin{table}[!t]
    \centering
    \caption{Impact of Perception Capability on Fast Reactive Navigator.}
    \renewcommand\arraystretch{1.2}
    \belowrulesep=0pt
    \aboverulesep=0pt
    \resizebox{0.4\textwidth}{!}{
    \begin{tabular}{c|c|ccc}
    \toprule
    Observation Source & $\phi_{iou}$ & SR & OSR & SPL \\
    \cmidrule{1-5}
    \multirow{5}{*}{Ideal Perception}
    & 0.05 & 35.9 & 42.5 & 31.0 \\
    & 0.20 & 31.2 & 37.9 & 27.5 \\
    & 0.40 & 32.9 & 39.2 & 29.0 \\
    & 0.60 & 31.6 & 39.2 & 27.6 \\
    & 0.80 & 36.0 & 41.8 & 31.7 \\
    \cmidrule(lr){1-5}
    \multirow{4}{*}{3D SAM Models}
    & 0.063 & 23.4 & 35.3 & 17.7 \\
    & 0.125 & 26.4 & 35.6 & 21.1 \\
    & 0.461 & 24.1 & 36.9 & 19.1 \\
    & 0.567 & 23.5 & 33.1 & 19.1 \\
    \bottomrule
    \end{tabular}}
    \label{tab:fast}
    \vspace{-0.6cm}
\end{table}

\subsection{Practical Perception Model Analysis}
From the ESAM variant results in Fig.\ref{fig:results}, navigation performance is consistently bounded by the corresponding statistical upper bound, supporting our simulated upper-bound analysis. Below, we further analyze the reasons for the gap between the ESAM results and the statistical simulation.

For the slow LLM planner, this gap is mainly attributed to false positives, whereas our simulation only models false negatives. Missing detections primarily result in incomplete scene information, which usually has limited influence on the reasoning process of LLM. In contrast, false detections introduce incorrect semantic cues, which can mislead the LLM into forming an incorrect understanding of the scene, or even misinterpreting the functional role of a space.

For the fast reactive navigator,our analysis shows that Through Skill is highly sensitive to object dimensionality $\mathbf{s}$. In the simulation, even when centroid drift is introduced, the relative dimensions $(w, h, d)$ remain unchanged, which still permits correct normal-vector estimation for traversable objects such as doors. However, practical perception models often suffer from aspect-ratio distortions. For example, the predicted thickness and width may become nearly identical or even be swapped. Such structural inaccuracies lead to false heading vectors in the Through Skill and ultimately cause navigation failure.

In addition, despite the large difference in vocabulary size across ESAM variants, the model with a vocabulary size of 20 achieves 33.7 in SR for the slow LLM planner and 23.4 for the fast reactive navigator, nearly matching models with a vocabulary size of 200 under the same setting. This suggests that a vocabulary size of 20 is sufficient to support successful navigation for R2R.

\section{CONCLUSIONS}
In this paper, we quantified the impact of 3D scene understanding on VLN performance and derived statistical upper bounds on SR for two core subsystems in a typical VLM-LLM navigation framework SFCo-Nav. We showed that navigation success rate exhibits a clear saturation effect as perception capability improves, and further validated these upper bounds with ESAM variants with different levels of perception performance. Our analysis also revealed that false positives and distorted bounding box aspect ratios are critical factors affecting navigation performance, and that a small set of core navigation-relevant object categories is sufficient for successful navigation.

\addtolength{\textheight}{-12cm}   





\normalem
\bibliographystyle{IEEEtran}
\bibliography{ref}

@article{SFcoNav,
  title={SFCo-Nav: Efficient Zero-Shot Visual Language Navigation via Collaboration of Slow LLM and Fast Attributed Graph Alignment},
  author={Xiong, Chaoran and Wei, Litao and Hu, Xinhao and Ma, Kehui and Xia, Ziyi and Jiang, Zixin and Sun, Zhen and Pei, Ling},
  journal={arXiv preprint arXiv:2603.01477},
  year={2026}
}

@article{SSM,
      author    = {C. Xiong and Y. Huang and F. Yu and C. Chen and Y. Wang and S. Xia and L. Pei},
      title     = {Sensing, Social, and Motion Intelligence in Embodied Navigation: A Comprehensive Survey},
      journal   = {arXiv preprint arXiv:2508.15354},
      year      = {2025},
}

@article{BLIP2,
  author    = {J. Li and D. Li and S. Savarese and S. Hoi},
  title     = {BLIP-2: Bootstrapping Language-Image Pre-Training with Frozen Image Encoders and Large Language Models},
  journal   = {arXiv preprint arXiv:2301.12597},
  year      = {2023},
}

@article{NavCoT,
  author    = {B. Lin and Y. Nie and Z. Wei and J. Chen and S. Ma and J. Han and H. Xu and X. Chang and X. Liang},
  title     = {NavCoT: Boosting LLM-Based Vision-and-Language Navigation via Learning Disentangled Reasoning},
  journal   = {IEEE Trans. Pattern Anal. Mach. Intell.},
  year      = {2025},
  volume    = {47},
  number    = {7},
  pages     = {5945--5957},
  doi       = {10.1109/TPAMI.2025.3554559}
}

@inproceedings{MapGPT,
  title     = {{MapGPT}: Map-Guided Prompting with Adaptive Path Planning for Vision-and-Language Navigation},
  author    = {J. Chen and B. Lin and R. Xu and Z. Chai and X. Liang and K.-Y. Wong},
  booktitle = {Proc. Annu. Meet. Assoc. Comput. Linguist. (ACL)},
  pages     = {9796--9810},
  year      = {2024},
  address   = {Bangkok, Thailand},
  publisher = {Assoc. Comput. Linguist.}
}

@inproceedings{NavGPT,
  title     = {{NavGPT}: Explicit Reasoning in Vision-and-Language Navigation with Large Language Models},
  author    = {G. Zhou and Y. Hong and Q. Wu},
  booktitle = {Proc. AAAI Conf. Artif. Intell. (AAAI)}, 
  pages     = {849--857},
  year      = {2024},
  publisher = {AAAI Press},
  doi       = {10.1609/aaai.v38i7.28597}
}

@inproceedings{OpenNav,
  title     = {Open-Nav: Exploring Zero-Shot Vision-and-Language Navigation in Continuous Environment with Open-Source LLMs},
  author    = {Y. Qiao and W. Lyu and H. Wang and Z. Wang and Z. Li and Y. Zhang and M. Tan and Q. Wu},
  booktitle = {Proc. IEEE Int. Conf. Robot. Autom. (ICRA)},
  year      = {2025}
}

@inproceedings{BLIP,
  title={Blip: Bootstrapping language-image pre-training for unified vision-language understanding and generation},
  author={Li, Junnan and Li, Dongxu and Xiong, Caiming and Hoi, Steven},
  booktitle={International conference on machine learning},
  pages={12888--12900},
  year={2022},
  organization={PMLR}
}

@inproceedings{Spatialbot,
  title={Spatialbot: Precise spatial understanding with vision language models},
  author={Cai, Wenxiao and Ponomarenko, Iaroslav and Yuan, Jianhao and Li, Xiaoqi and Yang, Wankou and Dong, Hao and Zhao, Bo},
  booktitle={Proc. IEEE Int. Conf. Robot. Autom. (ICRA)},
  pages={9490--9498},
  year={2025}
}

@inproceedings{RAM,
  title={Recognize anything: A strong image tagging model},
  author={Zhang, Youcai and Huang, Xinyu and Ma, Jinyu and Li, Zhaoyang and Luo, Zhaochuan and Xie, Yanchun and Qin, Yuzhuo and Luo, Tong and Li, Yaqian and Liu, Shilong and others},
  booktitle={Proc. IEEE/CVF Conf. Comput. Vis. Pattern Recognit. (CVPR)},
  pages={1724--1732},
  year={2024}
}

@inproceedings{Fast-RCNN,
  title={Fast r-cnn},
  author={Girshick, Ross},
  booktitle={Proc. IEEE Int. Conf. Comput. Vis. (ICCV)},
  pages={1440--1448},
  year={2015}
}

@article{esam, 
      title={EmbodiedSAM: Online Segment Any 3D Thing in Real Time}, 
      author={Xiuwei Xu and Huangxing Chen and Linqing Zhao and Ziwei Wang and Jie Zhou and Jiwen Lu},
      journal={arXiv preprint arXiv:2408.11811},
      year={2024}
}

@inproceedings{R2R,
  title     = {Vision-and-Language Navigation: Interpreting Visually-Grounded Navigation Instructions in Real Environments},
  author    = {P. Anderson and others},
  booktitle = {Proc. IEEE Conf. Comput. Vis. Pattern Recognit. (CVPR)},
  pages     = {3674--3683},
  year      = {2018}
}

@article{GPT,
    title     = {GPT-4 Technical Report},
    author    = {{OpenAI}},
    journal   = {arXiv preprint arXiv:2303.08774},
    year      = {2024},
    url       = {https://arxiv.org/abs/2303.08774}
}

@article{szot2021habitat,
  title     =     {Habitat 2.0: Training Home Assistants to Rearrange their Habitat},
  author    =     {Andrew Szot and others},
  journal   =     {arXiv preprint arXiv:2106.14405},
  year      =     {2021}
}

@inproceedings{scannet,
  title={Scannet: Richly-annotated 3d reconstructions of indoor scenes},
  author={Dai, Angela and Chang, Angel X and Savva, Manolis and Halber, Maciej and Funkhouser, Thomas and Nie{\ss}ner, Matthias},
  booktitle={Proc. IEEE Conf. Comput. Vis. Pattern Recognit. (CVPR)},
  pages={5828--5839},
  year={2017}
}

\end{document}